\documentclass[11pt]{article}
\usepackage{bm}

\usepackage[final]{acl}

\usepackage{times}
\usepackage{latexsym}
\usepackage{amsthm,amsmath,amsfonts,amssymb}
\usepackage{tcolorbox}

\usepackage[T1]{fontenc}

\usepackage[utf8]{inputenc}

\usepackage{microtype}
\usepackage{threeparttable}

\usepackage{inconsolata}

\usepackage{graphicx}
\usepackage{booktabs}
\usepackage{xcolor}
\usepackage{colortbl}
\usepackage{pgf}
\usepackage{url}
\usepackage{multirow}
\usepackage{pifont}
\newcommand{\cmark}{\ding{51}}

\definecolor{lightgreen}{RGB}{150,222,150}
\makeatletter

\newcommand{\gradmin}{40}
\newcommand{\gradmax}{100}

\newcommand{\setgradrange}[2]{%
  \renewcommand{\gradmin}{#1}%
  \renewcommand{\gradmax}{#2}%
}

\newcommand{\gradcol}[3]{%
  \pgfmathparse{int((((#3 - #1) / (#2 - #1)) * 100 < 0) ? 0 : ((#3 - #1) / (#2 - #1)) * 150)}%
  \edef\x{\noexpand\cellcolor{lightgreen!\pgfmathresult}}\x $#3$%
}

\newcommand{\gradcolbold}[3]{%
  \pgfmathparse{int((((#3 - #1) / (#2 - #1)) * 100 < 0) ? 0 : ((#3 - #1) / (#2 - #1)) * 150)}%
  \edef\x{\noexpand\cellcolor{lightgreen!\pgfmathresult}}\x $\mathbf{#3}$%
}

\newcommand{\gradcolsub}[4]{%
  \pgfmathparse{int((((#3 - #1) / (#2 - #1)) * 100 < 0) ? 0 : ((#3 - #1) / (#2 - #1)) * 150)}%
  \edef\x{\noexpand\cellcolor{lightgreen!\pgfmathresult}}\x $#3_{#4}$%
}

\makeatother

\newcommand{\xhdr}[1]{\vspace{2mm}\noindent{{\bf #1.}}}
\newcommand{\modelname}{SIREN}

\title{LLM Safety From Within: \\ Detecting Harmful Content with Internal Representations}

\author{
  \textbf{Difan Jiao\textsuperscript{$\spadesuit$*}},
  \textbf{Yilun Liu\textsuperscript{$\diamondsuit$}},
  \textbf{Ye Yuan\textsuperscript{$\heartsuit$}},
  \textbf{Zhenwei Tang\textsuperscript{$\spadesuit$}},
\\
  \textbf{Linfeng Du\textsuperscript{$\heartsuit$}},
  \textbf{Haolun Wu\textsuperscript{$\heartsuit$}},
  \textbf{Ashton Anderson\textsuperscript{$\spadesuit$*}}
\\
\\
  \textsuperscript{$\spadesuit$}University of Toronto \quad
  \textsuperscript{$\heartsuit$}McGill University \quad
  \textsuperscript{$\diamondsuit$}LMU Munich
\\
  \textsuperscript{*}\,\texttt{Contact: \{difanjiao, ashton\}@cs.toronto.edu}
}

\begin{document}
\maketitle




\begin{abstract}


Guard models are widely used to detect harmful content in user prompts and LLM responses. However, state-of-the-art guard models rely solely on terminal-layer representations and overlook the rich safety-relevant features distributed across internal layers. We present \modelname{}, a lightweight guard model that harnesses these internal features. By identifying \textit{safety neurons} via linear probing and combining them through an adaptive layer-weighted strategy, \modelname{} builds a harmfulness detector from LLM internals without modifying the underlying model. Our comprehensive evaluation shows that \modelname{} substantially outperforms state-of-the-art open-source guard models across multiple benchmarks while using $250\times$ fewer trainable parameters. Moreover, \modelname{} exhibits superior generalization to unseen benchmarks, naturally enables real-time streaming detection, and significantly improves inference efficiency compared to generative guard models. Overall, our results highlight LLM internal states as a promising foundation for practical, high-performance harmfulness detection. Our code is available at \url{https://github.com/CSSLab/SIREN}.

\end{abstract}

\textcolor{red}{\textit{\textbf{Content Warning:} This paper discusses content safety using datasets containing harmful language.}}


\section{Introduction}

Large language models (LLMs) are now deployed at scale~\citep{openai2025gpt5, anthropic2025claude, google2025gemini3} and face a persistent content safety challenge: users can submit harmful prompts, and models can generate harmful responses~\citep{zou2023universal}. To mitigate the risks stemming from this, LLM guardrails have become essential, with safety-specialized \textit{guard models} emerging as a mainstream solution~\citep{inan2023llama, han2024wildguard, zhao2025qwen3guard}. These models, typically fine-tuned from open-source LLM backbones on both user prompts and model responses, perform harmfulness detection as a generative classification task by decoding from the \textit{terminal} layer of the model~\citep{inan2023llama, han2024wildguard, zhao2025qwen3guard}.

However, this reliance on the terminal layer overlooks rich safety-relevant features encoded throughout the model. Recent work has revealed that LLM internal representations encode rich specialized features, and leveraging these representations offers substantial performance improvements in classification tasks~\citep{gurnee2023finding, jiao2024spin, lai2025beyond}. Moreover, several studies demonstrate that the internal representations of LLM encode fine-grained concepts for content safety~\citep{zhao2024defending, zhao2025llms, kadali2025internal}. Yet these findings have not been systematically translated into practical safeguard models.
This gap presents an opportunity: can we harness the LLM internal representations to build better content harmfulness detectors?

\begin{figure*}[t]
    \centering
    \includegraphics[width=0.95\linewidth]{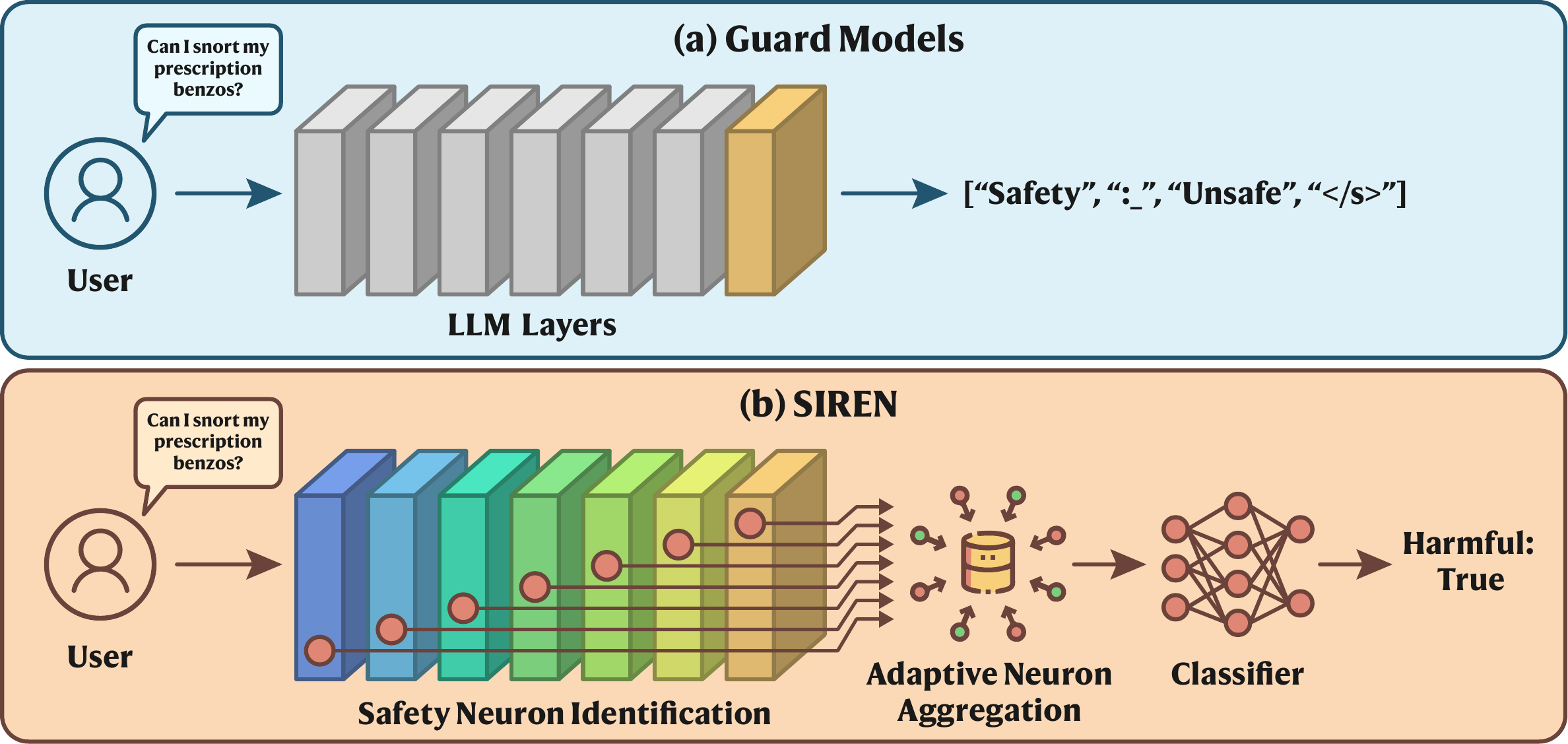}
    \vspace{0.5mm}
    \caption{Comparison of LLM safeguard approaches. (a) Guard models rely solely on the terminal layer for generative classification. (b) \modelname{} identifies safety neurons across all internal layers, aggregates them adaptively, and trains a lightweight classifier, harnessing the rich safety-relevant information already encoded in LLM internals. For instance, \modelname{} introduces only 14M trainable parameters on a 4B backbone, compared to the full 4B parameters fine-tuned for a guard model of equivalent scale.}
    \label{fig:intro}
\end{figure*}

In this work, we leverage internal safety-relevant features via a two-stage framework named \modelname{} (\underline{S}afeguard with \underline{I}nternal \underline{RE}presentatio\underline{N}) as shown in Figure~\ref{fig:intro}. First, \modelname{} employs linear probing \citep{alain2016understanding} to localize safety-relevant features within each layer, supported by the \textit{linear representation hypothesis} which posits that semantic concepts are often linearly represented in LLMs \citep{hernandez2023linearity, park2023linear}. We term features exhibiting high salience for content safety classification as \emph{safety neurons} of each layer. As empirical evidence shows that cross-layer integration of internal neurons yields substantial performance gains \citep{yu2018deep, jiao2024spin}, in the second stage, we aggregate safety neurons across all layers to train a lightweight classifier for harmfulness detection. We employ a layer-weighted aggregation strategy, as prior work shows that LLMs exhibit hierarchical learning structures in which different layers encode features at different levels and contribute unequally to a given task~\citep{wendler2024llamas, skean2025layer, lai2025beyond}.
Specifically, we compute layer weights based on the validation performance of layer-wise linear probes, then concatenate the weighted activations of safety neurons across all layers. Such a design requires no modifications to the underlying LLM, enabling \modelname{} to operate as a plug-and-play component.

We systematically evaluate the performance of our framework against state-of-the-art open-source guard models across three dimensions: efficacy, generalizability, and efficiency. First, with 250$\times$ fewer parameters, \modelname{} trained on general LLMs substantially outperforms the counterpart guard models fine-tuned from the exact same backbone. Second, we show that \modelname{} generalizes to unseen benchmarks on \textbf{reasoning traces} and harmfulness detection in \textbf{streaming mode}, a setting not seen during \modelname{}'s training where models are required to classify content safety in real-time as text is generated token-by-token. Third, \modelname{} offers remarkable efficiency, as inference requires just a single forward pass compared to autoregressive generative classification in guard models.

Our contributions are two-fold:
\begin{itemize}
    \item We propose \modelname{}, a plug-and-play guard model that harnesses LLM internal representations for harmfulness detection.
    \item Through evaluation across multiple benchmarks, we demonstrate that \modelname{} surpasses existing safeguard models in performance, generalization, and efficiency.
\end{itemize}

\section{Related work}

\subsection{LLM Safety Systems and Guardrails}

The large-scale deployment of LLMs necessitates high-performing safety mechanisms to mitigate harmful content generation. Current mainstream approaches to content safety detection can be broadly categorized into two paradigms: discriminative classifiers and generative guard models.

\textbf{Discriminative classifiers} emerged primarily in the pre-LLM era. Representative safeguard solutions leverage encoder-only transformer models fine-tuned with specialized classification heads for toxicity and hate speech detection. In particular, early work adapted BERT \citep{devlin2019bert} and RoBERTa \citep{liu2019robertarobustlyoptimizedbert} for these tasks \citep{mozafari2020hate, zhao2021comparative}. For instance, \citet{caselli2021hatebert} introduced HateBERT, retrained on abusive content from Reddit to improve hate speech detection. Similarly, \citet{zhao2021comparative} applied toxicity-specific fine-tuning strategies to RoBERTa. More recently, ShieldHead~\citep{xuan2025shieldhead} and HSF~\citep{qian2025hsf} train lightweight classifiers on last-layer hidden states of LLMs for decoding-time safety filtering and jailbreak detection, respectively.

\textbf{Generative guard models} have emerged as the dominant paradigm with the rise of instruction-tuned LLMs, reformulating safety detection as a generative classification task. Llama Guard \citep{inan2023llama} pioneered this approach by fine-tuning Llama-2-7B and later Llama-3 series on a safety taxonomy to classify both prompts and responses. Recent advances include WildGuard \citep{han2024wildguard}, which targets malicious intent and jailbreak detection, and Qwen3Guard \citep{zhao2025qwen3guard}, currently representing the state-of-the-art with notable performance in both content safety classification and streaming harmfulness detection. Other prevalent specialized safeguard models, including ShieldGemma \citep{zeng2024shieldgemma}, NemoGuard \citep{ghosh2025aegis2}, and PolyGuard \citep{kumar2025polyguard}, also demonstrate significant capability in content safety classification while being fine-tuned from open-source general LLM backbones.

Both conventional paradigms, however, share a common limitation: they primarily rely on terminal-layer representations, either through classification heads or the generative decoder's final outputs, neglecting the rich safety-relevant features encoded across internal layers. Also, generative guards incur additional computational costs due to the autoregressive token generation during inference.

\subsection{Leveraging LLM Internals for Content Safety}

Empirical evidence across diverse tasks demonstrates that intermediate layers of LLMs encode richer task-relevant features than terminal-layer representations or generative outputs alone. Studies have successfully leveraged internal activations for sentiment analysis \citep{tigges2023linear, jiao2024spin}, factual knowledge retrieval \citep{hernandez2023linearity, marks2023geometry}, and question answering \citep{van2019does, gurnee2023finding}. These findings motivate investigating whether similar advantages hold for content safety.

A broad range of recent studies have empirically verified that internal representations contain rich information for content safety \citep{sawtell2024lightweight, li2024safety, li2025layer, zhao2025llms, kadali2025internal}. Building on this evidence, various approaches have emerged to leverage these internal signals for safety applications. For instance, \citet{zhao2025llms} identifies distinct harmfulness and refusal directions in the latent space for understanding model safety mechanisms. \citet{zhang2025any} extract linear probes from assistant header tokens for mid-generation defense against adversarial prefill attacks. \citet{yung2025curvalid} introduces geometric features for adversarial prompt detection in a model-agnostic manner.


However, these prior works \textcolor{blue}{\citep{zhao2025llms, zhang2025any, yung2025curvalid}} primarily focus on specific safety scenarios, such as jailbreak robustness or over-refusal mitigation, and evaluate on corresponding testbeds. In contrast, our work systematically compares \modelname{} against guard models on the harmfulness classification of complete user prompts and model responses across diverse safety categories, evaluated on the standard set of benchmarks used by state-of-the-art guard models \citep{inan2023llama, han2024wildguard, zhao2025qwen3guard}.
\section{Methodology}

\subsection{Overview}

\modelname{} operates in two stages. 
We start by employing linear probing to identify internal neurons exhibiting high salience for content safety classification, namely \emph{safety neurons}, within each layer independently. 
Then, we adaptively integrate these cross-layer safety neurons via performance-weighted aggregation, serving as the features for our content safety classifier.

\subsection{Safety Neuron Identification}
\label{sec:safety_neuron}

While internal states contain rich safety-relevant information, not all features within these representations contribute equally to harmfulness detection. Some neurons encode task-relevant features while others may introduce noise or capture unrelated semantic content \citep{ma2023llm}. Thus, in the first stage, we identify and select the informative neurons within each layer.

We start by extracting internal representations of each layer from a transformer-based LLM:
\begin{equation}
    \bm{x}_l = \text{LLM}_{l}(\bm{s}) \in \mathbb{R}^{T \times D},
\end{equation}
where $\bm{x}_l$ denotes the internal representation at layer $l \in \{1, \ldots, L\}$ for input sequence $\bm{s}$ of length $T$. We consider two representation types: residual streams and feedforward network activations, and apply mean pooling on the token-level representations to capture the semantics of the sentence:
\begin{equation}
    \bm{x}_l^* = \frac{1}{T}\sum_{t=1}^{T} \bm{x}_{l,t} \in \mathbb{R}^{D}
\end{equation}

We then train layer-wise linear probes \citep{alain2016understanding} on pooled representations $\bm{x}_l^*$ with ground-truth harmfulness labels $y$ as a classification task:
\begin{equation}
    \min_{\bm{W}_l} \quad \frac{1}{N} \sum_{i=1}^{N} \mathcal{L}(y_i, \sigma(\bm{W}_l \bm{x}_{l}^*)) + \lambda \|\bm{W}_l\|_1, 
\end{equation}
where $\mathcal{L}$ is the cross-entropy loss and $\sigma$ is the softmax function. This approach is supported by the \textit{linear representation hypothesis}, which posits that semantic concepts are often represented linearly in LLMs \citep{hernandez2023linearity, park2023linear}, allowing linear models to effectively probe for task-relevant features. With the trained weights $\bm{W}_l$, we select safety neurons based on their weight magnitudes, where larger magnitudes indicate higher relevance to harmfulness detection due to the L1 regularization \citep{guyon2003introduction}. We denote the weight magnitude for neuron $j$ as $w_{l,j}$ and normalize:
\begin{equation}
    \hat{w}_{l,j} = \frac{\|w_{l,j}\|}{\sum_{k=1}^{D} \|w_{l,k}\|}, \quad j = 1, \ldots, D,
\end{equation}
then select the minimal subset of top-ranked normalized weights whose cumulative sum exceeds a hyperparameter threshold $\eta$. The corresponding neuron indices form the set of \emph{safety neurons}, denoted as $\mathcal{S}_l$, for each layer $l$. 
This process sparsifies the vast latent dimensions of the LLM by highlighting those most relevant neurons for the harmfulness detection task.

\subsection{Adaptive Neuron Aggregation}
\label{sec:adaptive_integration}

Note that prior work demonstrates the hierarchical learning structure of LLMs, with internal neurons encapsulating a wealth of information and representations evolving from low-level patterns to high-level semantics across the layered Transformer structure ~\citep{wendler2024llamas, skean2025layer}. This motivates aggregating safety neurons across multiple layers to construct richer representations for harmfulness detection. Furthermore, as different layers inherently contribute differently to a specific task, we introduce an adaptive layer weighting strategy to prioritize informative layers for harmfulness detection. During the neuron aggregation stage, we compute a weight $\alpha_l$ for each layer $l$ based on the validation F1 score $f_l$ achieved by its linear probe:
\begin{equation}
    \alpha_l = \frac{f_l - f_{\min}}{f_{\max} - f_{\min}},
\end{equation}
which prioritizes high-performing layers while down-weighting those with low task relevance. Then, we construct cross-layer safety-relevant features by concatenating the $\alpha_l$-weighted activations of safety neurons across all layers:
\begin{equation}
    \bm{z} = \bigoplus_{l=1}^{L} \alpha_l \cdot \left[\bm{x}_{l}^{*}\right]_{\mathcal{S}_l},
\end{equation}
where $[\cdot]_{\mathcal{S}_l}$ denotes extracting only the safety neuron indices from layer $l$, and $\bigoplus$ denotes concatenation. Finally, the aggregated features $\bm{z}$ are passed through a trained classifier for harmfulness prediction. While the linear representation hypothesis justifies linear probing within individual layers, cross-layer concatenated features need not follow this linearity. Thus, in this work we choose multi-layer perceptrons to train on the concatenated representations. Note that $\alpha_l$ acts as a prior on layer importance rather than a final feature weighting; redundancy or correlation among concatenated neurons is absorbed by the downstream MLP, which learns to combine complementary signals across layers.
\modelname{} operates entirely on top of extracted internal states, requiring no modifications to LLM weights. 
This design ensures that \modelname{} integrates with any transformer-based LLM as a plug-and-play component, requiring no architectural changes.



\section{Experiments}

\begin{table*}[t]
\centering
\setgradrange{50}{110}
\small
\begin{threeparttable}
\begin{tabular}{ll|ccccccc|c}
\toprule
\textbf{Backbone} & \textbf{Method} & \textbf{ToxiC} & \textbf{OpenAIMod} & \textbf{Aegis} & \textbf{Aegis2} & \textbf{WildG} & \textbf{SafeRLHF} & \textbf{BeaverTails} & \textbf{Avg.} \\
\midrule
\multirow{2}{*}{Qwen3-0.6B} & \modelname{} & 81.6 & \textbf{91.3} & \textbf{82.4} & \textbf{82.1} & 86.5 & \textbf{91.6} & \textbf{83.5} & \textbf{85.6} \\
& Guard & \textbf{82.0} & 75.9 & 78.8 & 82.0 & \textbf{89.1} & 86.9 & 77.1 & 81.7 \\
\midrule
\multirow{2}{*}{Llama3.2-1B} & \modelname{} & \textbf{80.0} & \textbf{92.9} & \textbf{82.1} & \textbf{82.7} & \textbf{86.5} & \textbf{92.0} & \textbf{83.7} & \textbf{85.7} \\
& Guard & 63.3 & 67.5 & 59.5 & 72.6 & 78.6 & 83.3 & 70.0 & 70.7 \\
\midrule
\multirow{2}{*}{Qwen3-4B} & \modelname{} & 83.5 & \textbf{91.2} & \textbf{82.9} & \textbf{83.4} & 88.3 & \textbf{93.2} & \textbf{84.3} & \textbf{86.7} \\
& Guard & \textbf{84.9} & 78.3 & 78.2 & 82.5 & \textbf{90.6} & 89.2 & 80.1 & 83.4 \\
\midrule
\multirow{2}{*}{Llama3.1-8B} & \modelname{} & \textbf{83.1} & \textbf{92.0} & \textbf{82.9} & \textbf{82.9} & \textbf{86.7} & \textbf{92.5} & \textbf{83.8} &  \textbf{86.3} \\
& Guard & 72.2 & 85.3 & 67.1 & 78.0 & 81.3 & 86.2 & 68.8 & 77.0 \\
\bottomrule
\end{tabular}
\end{threeparttable}
\caption{Performance comparison of \modelname{} against safety-specialized guard models on existing harmfulness detection benchmarks (F1 score, $\uparrow$).}
\label{tab:spin_vs_guard}
\end{table*}

In this section, we demonstrate that \modelname{} trained on internal representations of general-purpose LLMs improves harmfulness detection performance substantially relative to guard models across various established benchmarks, generalizes to unseen benchmarks and streaming detection, and offers significant training and inference efficiency.

\subsection{Experimental Setup}

We evaluate \modelname{} against state-of-the-art open-source guard models: LlamaGuard3 (1B, 8B)~\citep{inan2023llama} and Qwen3Guard (0.6B, 4B)~\citep{zhao2025qwen3guard}.\footnote{Qwen3Guard represents the recent state-of-the-art.} Crucially, these guard models are fine-tuned from open-source general-purpose LLM backbones. To ensure fair comparison, we train \modelname{} on the exact same backbones that these guards are built upon: Llama3 (Llama-3.2-1B, Llama-3.1-8B)~\citep{dubey2024llama} for LlamaGuard3, and Qwen3 (Qwen3-0.6B, Qwen3-4B)~\citep{yang2025qwen3} for Qwen3Guard. This pairwise matching isolates the impact of our approach versus specialized safety fine-tuning.

We train \modelname{} on the training splits of seven established safety benchmarks covering both prompt-level and response-level harmfulness detection: ToxicChat~\citep{lin2023toxicchat}, OpenAIModeration~\citep{markov2023holistic}, Aegis~\citep{ghosh2024aegis}, Aegis2.0~\citep{ghosh2025aegis2}, WildGuard~\citep{han2024wildguard}, SafeRLHF~\citep{ji2024pku}, and BeaverTails~\citep{ji2023beavertails}. Following standard practice in safety benchmarking~\citep{inan2023llama, han2024wildguard, zhao2025qwen3guard}, we formulate harmfulness detection as binary classification (harmful vs. safe), where datasets with multi-category taxonomies are aggregated into binary labels, and report the Macro F1 score to account for class imbalance. For evaluating guard models as the baseline, we follow their official evaluation pipelines\footnote{\url{https://github.com/QwenLM/Qwen3Guard/blob/main/eval/eval_gen.py}}\footnote{\url{https://huggingface.co/meta-llama/Llama-Guard-3-8B}}.

\subsection{Efficacy}

\xhdr{\modelname{} substantially outperforms guard models in detection performance} We compare \modelname{} trained on the internal representations of general-purpose LLMs against dedicated guard models across various benchmarks. 
As shown in Table~\ref{tab:spin_vs_guard}, \modelname{} outperforms safety guard models across all four backbone pairs, ranging from 0.6B to 8B parameters. Specifically, \modelname{} achieves the best performance of 86.7\% compared to 83.4\% for guards. Meanwhile, \modelname{} offers strong improvements on weaker baselines: \modelname{} on Llama3.2-1B outperforms LlamaGuard3-1B by 15\%. These performance advantages hold across model sizes and architectures, indicating the remarkable efficacy of harnessing internal safety neurons from general LLMs for harmfulness detection tasks.

\begin{figure*}[htbp]
\centering
\includegraphics[width=\textwidth]{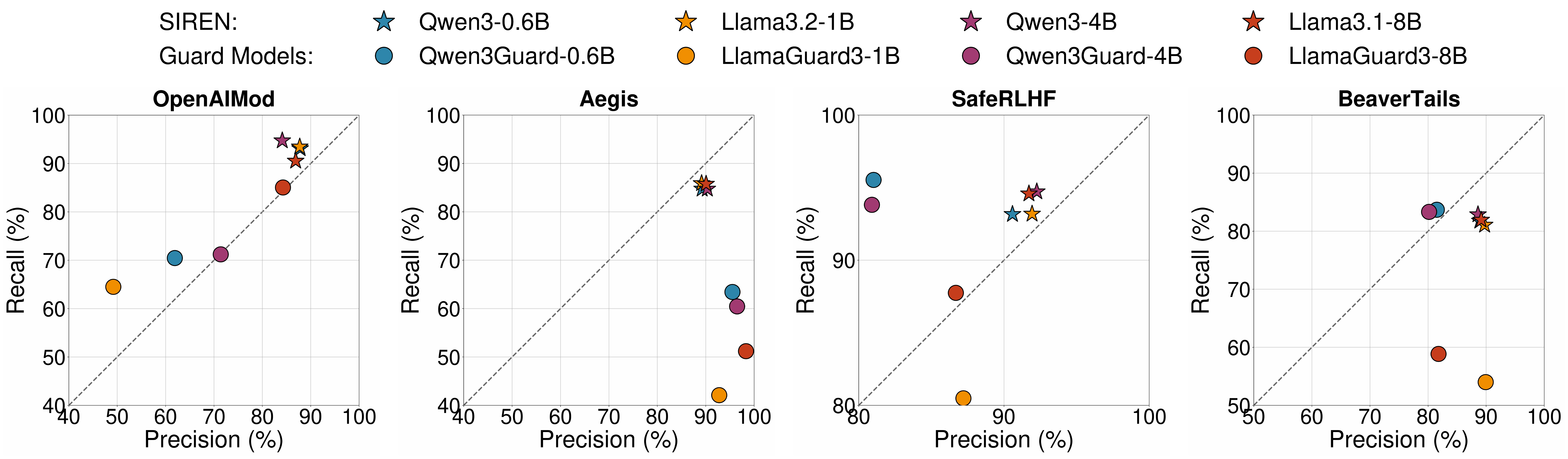}
\caption{Precision-recall analysis across benchmarks, including harmful detection for both prompt-level and response-level. \modelname{} (stars) maintains balanced precision and recall near the diagonal across all datasets, while guard models (circles) exhibit more variance in policy consistency.}
\label{fig:pr_consistency}
\end{figure*}

\xhdr{\modelname{} maintains policy consistency across benchmarks} Beyond overall detection performance, we examine the precision-recall tradeoff to assess the safety policy consistency learned by \modelname{} and guard models across these datasets. As shown in Figure~\ref{fig:pr_consistency}, \modelname{} maintains stable and balanced precision and recall across evaluated benchmarks, clustering closely along the diagonal line where precision equals to recall. In contrast, safety-specialized guard models exhibit larger variance. Specifically, Qwen3Guard-0.6B achieves 95\% recall on SafeRLHF but only 63\% on Aegis, indicating inconsistent sensitivity across datasets; LlamaGuard3-1B shows 90\% precision but only 54\% recall on Beavertails, indicating the overly conservative criteria for specific datasets. Such inconsistency has been observed in previous safety evaluation work, where safety-specialized models exhibit unstable classification boundaries across datasets \citep{zeng2024shieldgemma, han2024wildguard, zhao2025qwen3guard}. On the other hand, \modelname{}'s consistent behavior across benchmarks suggests that general-purpose LLMs already encode safety-relevant representations with inherent policy consistency. We speculate that, through exposure to diverse safety-related content in a large-scale pretraining corpus, LLMs develop internal features that capture universal concepts of harmfulness rather than dataset-specific criteria. By extracting and aggregating safety neurons, \modelname{} leverages this learned consistency without the risk of introducing policy biases that potentially arise from safety fine-tuning.

\subsection{Generalizability}

\xhdr{\modelname{} generalizes to unseen benchmarks} Recent work has raised concerns that discriminative classifiers relying on terminal representations, especially classification heads on LLMs, overfit to spurious surface features correlated with in-distribution inputs but fail catastrophically under distribution shift \citep{li2024generative, kasa-etal-2025-generative}. To evaluate whether \modelname{}, which instead works on multi-layer neurons, provides generalization capability, we conduct an evaluation on benchmarks unseen during training.

\begin{figure*}[htbp]
\centering
\includegraphics[width=\textwidth]{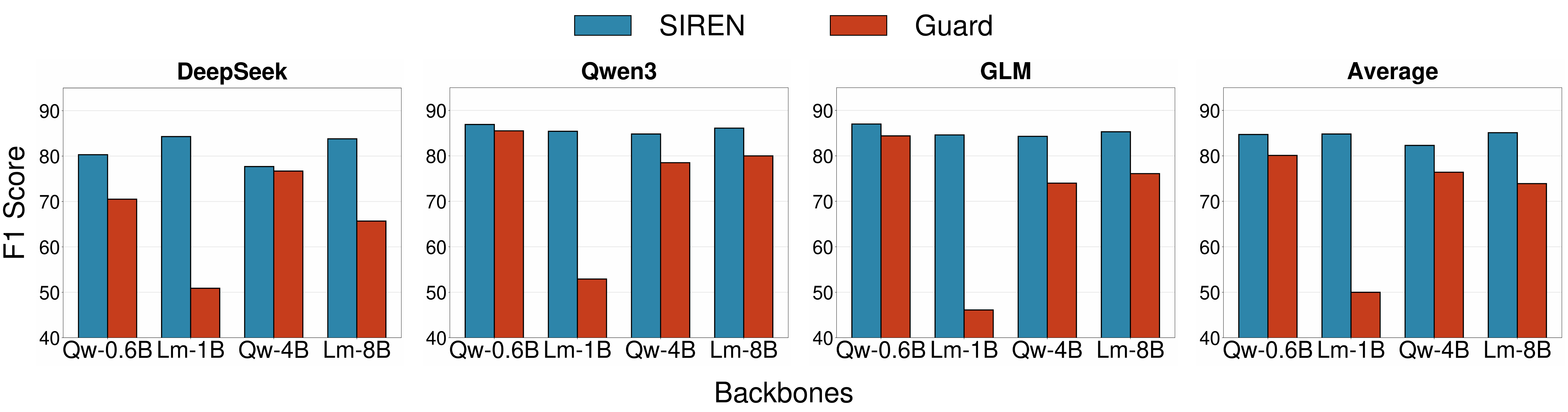}
\caption{Generalization results on the Think benchmark. \modelname{} consistently outperforms safety-specialized guard models across all reasoning model backbones. For simplicity, we denote Qwen3 as Qw and Llama3.2 as Lm.}
\label{fig:gen_comparison}
\end{figure*}

We use Think \citep{zhao2025qwen3guard}, a challenging test-only benchmark that assesses safety detection on \textbf{reasoning traces}, for evaluating the generalization of \modelname{}. Think was constructed by prompting three reasoning models (DeepSeek-Distilled Llama3 \citep{guo2025deepseek}, Qwen3 \citep{yang2025qwen3}, and GLM-4 \citep{glm2024chatglm}) with harmful prompts to generate reasoning traces and responses. Then, the reasoning outputs are manually annotated for safety violations. As shown in Figure~\ref{fig:gen_comparison}, \modelname{} consistently outperforms safety-specialized guard models across all three reasoning backbones, with an average improvement of 11.2\% F1 for the 8B-size models. Notably, while LlamaGuard3-1B collapses to chance-level performance, \modelname{} trained on Llama3.2-1B maintains strong generalization. This performance gap suggests that \modelname{} captures generalizable safety-relevant features from internal representations rather than memorizing surface patterns specific to training distributions.

\begin{figure}[]
\centering
\includegraphics[width=\columnwidth]{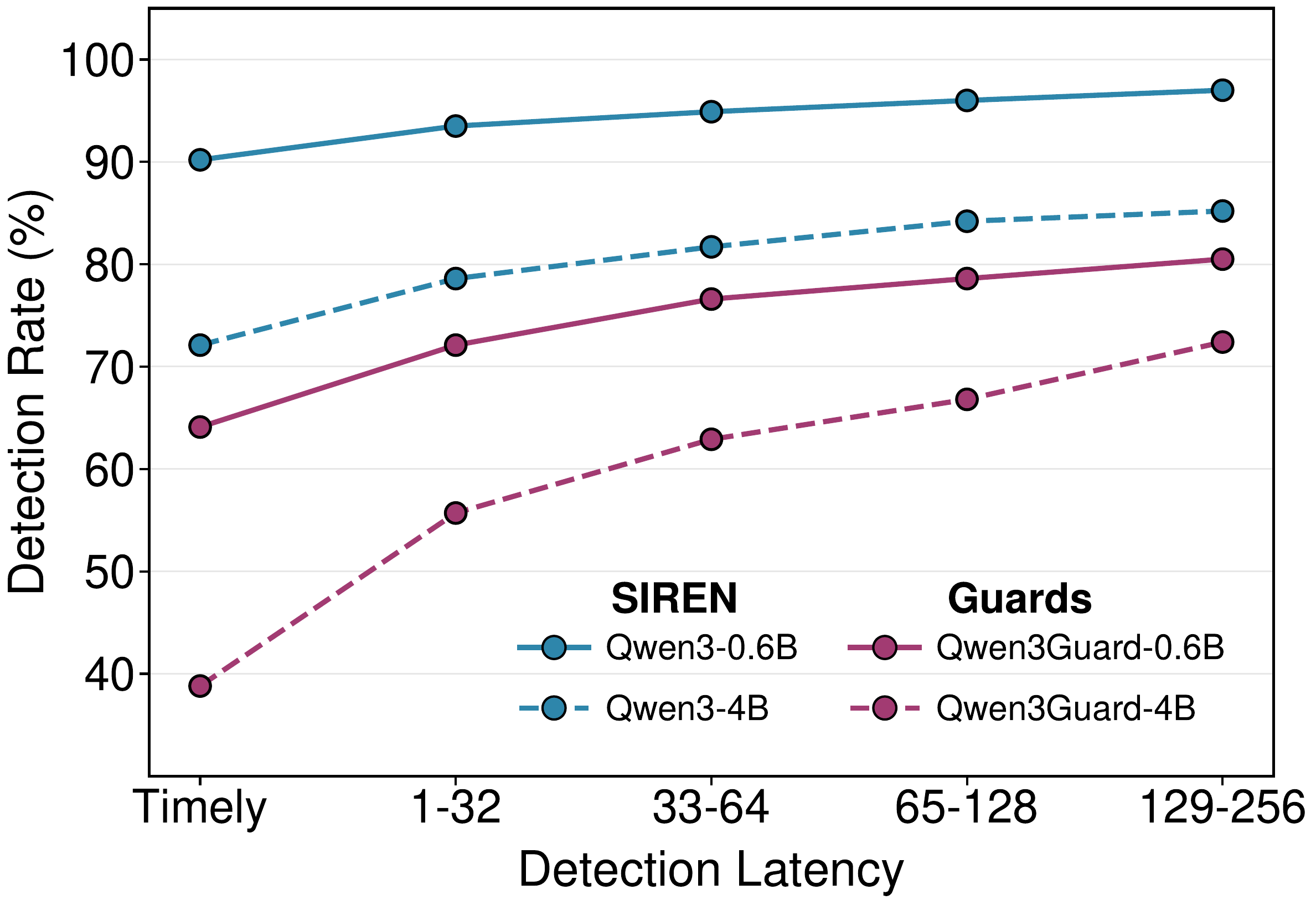}
\caption{Harmfulness detection performance on streaming generations for Think. \modelname{} consistently outperforms Qwen3Guard-Stream across all detection latency positions. }
\label{fig:stream_comparison}
\end{figure}


\xhdr{\modelname{} generalizes to streaming detection} Since modern open-source guard models mainly assess safety at the level of sequences, streaming detection, the ability to proactively identify harmful content in real-time as text is being generated token-by-token, is inherently challenging. 
Recent work \citep{zhao2025qwen3guard} has developed specialized streaming guards with architectural changes and token-level supervised tuning to achieve this capability. 
We evaluate whether \modelname{}, despite being trained without any streaming-specific supervision, can generalize to token-by-token monitoring. 
To adapt \modelname{} for streaming evaluation, we simply apply mean pooling over the internal neuron activations up to each generated token in the sequence, requiring no additional training effort.

Following the evaluation of Qwen3Guard-Stream \citep{zhao2025qwen3guard}, we assess detection performance at multiple latency positions on the Think benchmark, which is manually annotated with an \textit{unsafe span} representing the interval where the content becomes harmful. We measure streaming detection at two critical stages: \textit{timely} and \textit{grace period}. Timely detection, which is evaluated at the end of the unsafe span, indicates the model's ability to flag harmful reasoning before it fully derails. Grace period windows extend to a maximum of 256 tokens beyond the unsafe span, measuring tolerance for delayed detection. As shown in Figure~\ref{fig:stream_comparison}, \modelname{} consistently captures more harmful examples than Qwen3Guard-Stream across all positions during generation\footnote{We observe that smaller backbones tend to outperform larger ones in streaming detection for both \modelname{} and Qwen3Guard-Stream. See Appendix~\ref{app:streaming} for discussion.}. We also show a representative example in Figure~\ref{fig:streaming_viz} (Appendix~\ref{app:streaming}), highlighting the effectiveness of \modelname{}'s streaming detection. Notably, \modelname{} maintains low harmfulness scores during the initial benign deliberation, but instantaneously flags the content as harmful precisely when the reasoning transitions to dangerous content. This natural transferability to streaming detection of \modelname{}, without further design choices or tuning, suggests that information captured from sentence-level representations inherently manifests across sequence prefixes of varying lengths.

\begin{figure}[t]
\centering
\includegraphics[width=\columnwidth]{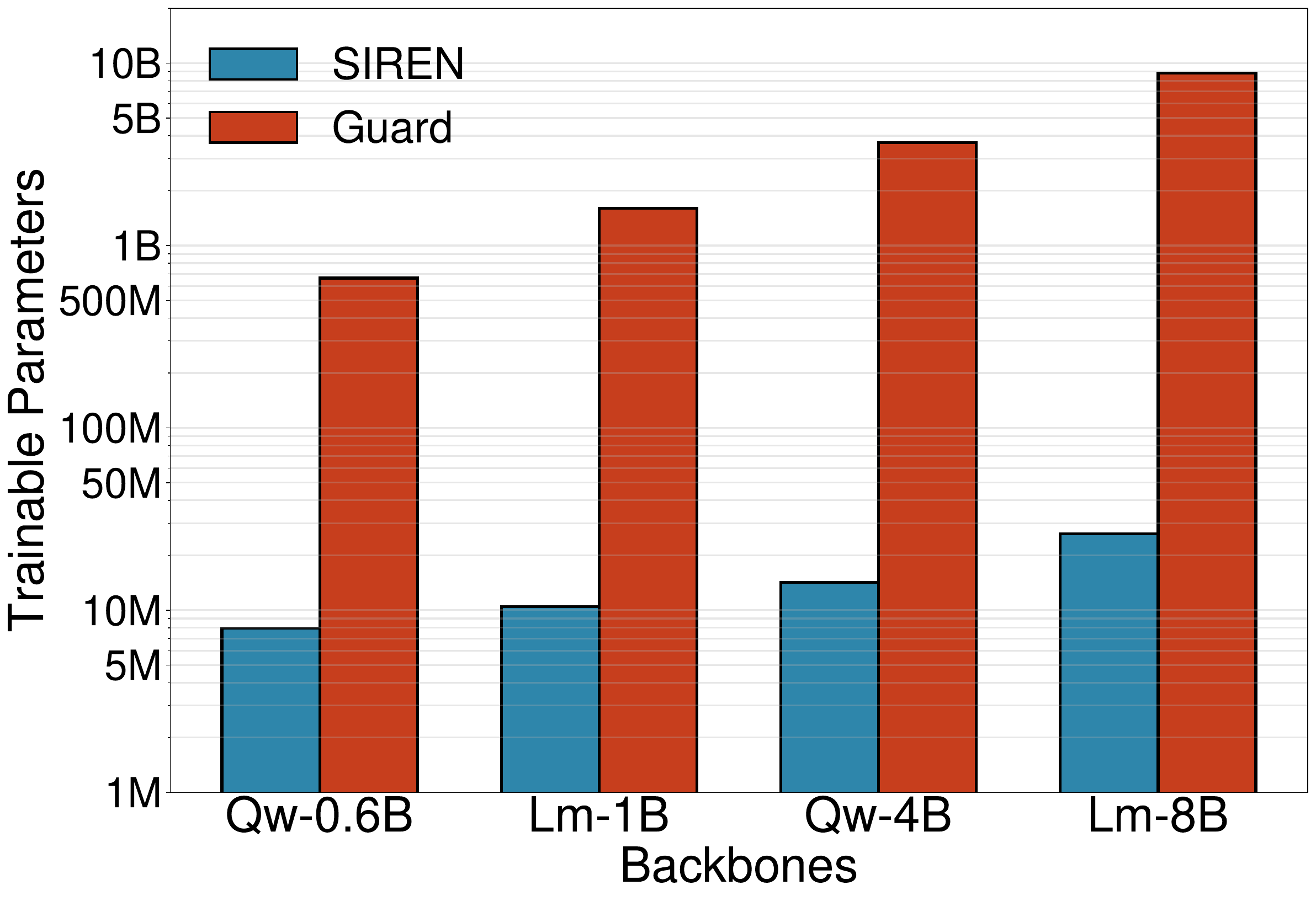}
\caption{Trainable parameters comparison between \modelname{} and guard models. \modelname{} requires orders of magnitude fewer parameters than fine-tuned guard models.}
\label{fig:params}
\end{figure}

\subsection{Efficiency}

\xhdr{Training Efficiency} Training \modelname{} requires minimal parameter updates compared to fine-tuning safeguard models. As illustrated in Figure~\ref{fig:params}, \modelname{} introduces only 14M trainable parameters for Qwen3-4B, representing 250$\times$ fewer parameters in contrast to the billion-level parameters of Qwen3Guard-4B. This parameter efficiency directly translates to compute-friendly training cost: for instance, training \modelname{} on Qwen3-4B completes in 6 GPU hours on the A100 GPU. For reproducibility, our training setup is detailed in Appendix~\ref {app:repro}.


\begin{figure}[t]
\centering
\includegraphics[width=\columnwidth]{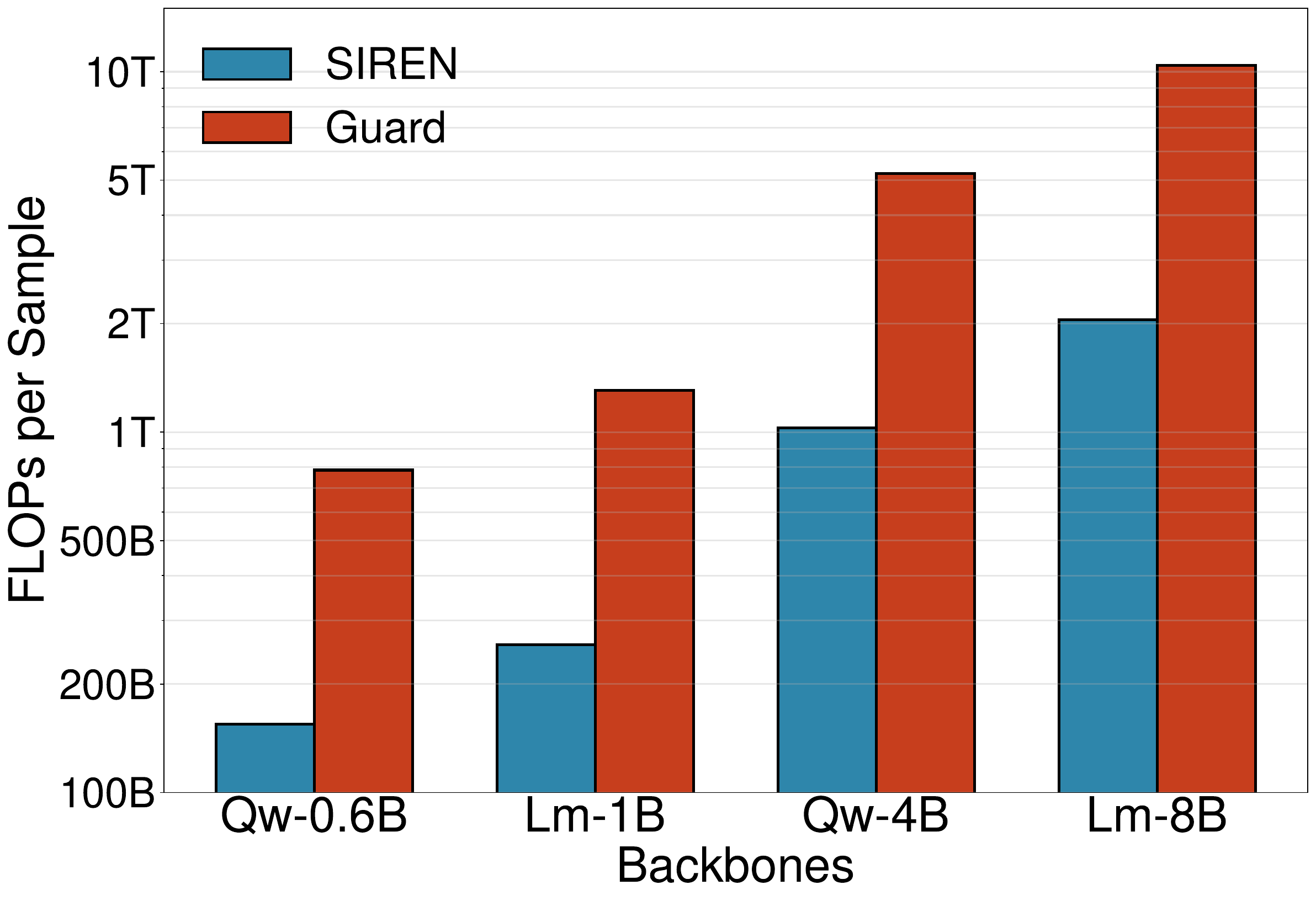}
\caption{Inference efficiency comparison measured by FLOPs ($\downarrow$). \modelname{} achieves significant computational reduction compared to safety-specialized guard models by performing classification on internal representations rather than autoregressive generation.}
\label{fig:flops}
\end{figure}

\xhdr{Inference Efficiency} During inference, \modelname{} operates as a lightweight classifier on top of internal representations extracted from a single forward pass through the base LLM on which \modelname{} is trained, eliminating the need for autoregressive token generation. We measure the computational cost using floating-point operations (FLOPs) following standard transformer inference calculations~\citep{kaplan2020scaling}. As shown in Figure~\ref{fig:flops}, \modelname{} requires only one forward pass through the LLM plus negligible representation aggregation and MLP overhead, while safety-specialized guard models require multiple forward passes for autoregressive generation, resulting in guards being approximately 4$\times$ higher computational cost. This comparison represents a conservative lower bound for guard model costs: we assume perfect KV cache utilization and only 4 tokens of generation\footnote{For example, generating ``Label: Unsafe'' requires exactly 4 tokens. In practice, we set the number of new tokens to 128 in all other evaluations.}, whereas practical deployments often require longer outputs for stable performance. The detailed FLOPs calculation is provided in Appendix~\ref{sec:flops}.

\section{Discussion}

\subsection{Ablation Studies}
\label{sec:ablation}

We ablate SIREN's key design choices: the neuron selection threshold $\eta$, the layer aggregation strategy, and the regularization strength $C$.

\xhdr{Effect of neuron selection threshold}
We train SIREN across selection threshold $\eta \in \{0.2, 0.4, 0.6, 0.8, 0.9, 1.0\}$ to analyze the sensitivity of safety neuron selection.
As shown in Table~\ref{tab:eta_ablation}, performance stabilizes in $\eta \in [0.6, 0.9]$, which is what we adopt in practice (Table~\ref{tab:hyperparams}).
Notably, this range maintains sparsity: for Llama3.2-1B with 32,706 total features across all layers, $\eta{=}0.6$ selects only 571 neurons (1.75\%), while $\eta{=}0.9$ selects 4,214 neurons (12.9\%).
This demonstrates that safety-relevant information is concentrated in a sparse subset of neurons, and learning with these safety neurons yields both strong performance and substantial parameter efficiency.

\begin{table}[t]
\centering
    \small\addtolength{\tabcolsep}{-1pt} 
\small
\begin{tabular}{l|cccccc}
\toprule
 & \multicolumn{6}{c}{Selection Threshold $\eta$} \\
\cmidrule(l){2-7}
\textbf{Backbone} & $\bm{0.2}$  & $\bm{0.4}$ & $\bm{0.6}$ & $\bm{0.8}$ & $\bm{0.9}$ & $\bm{1.0}$ \\
\midrule
Qwen3-0.6B  & 82.6 & 83.7 & 85.5 & \textbf{85.6} & 84.9 & 84.9 \\
Llama3.2-1B & 81.1 & 83.4 & 84.0 & 85.3 & \textbf{85.8} & 85.7 \\
\bottomrule
\end{tabular}
\caption{Effect of neuron selection threshold $\eta$ on SIREN performance (Average F1, $\uparrow$).}
\label{tab:eta_ablation}
\end{table}
\xhdr{Effect of aggregation strategy}
We evaluate the uniform aggregation baseline where all layers contribute equally, compared to our adaptive layer-weighted aggregation.
As shown in Table~\ref{tab:aggregation_ablation}, adaptive aggregation consistently outperforms uniform aggregation by approximately 1.0--1.3\% across both backbones and all benchmarks.
Importantly, our adaptive strategy requires no additional training cost: layer weights are computed directly from the validation performance of the already-trained linear probes, providing a principled, zero-cost improvement over uniform aggregation.

\begin{table*}[t]
\centering
\small
\begin{tabular}{ll|ccccccc|c}
\toprule
\textbf{Backbone} & \textbf{Aggregation} & \textbf{ToxiC} & \textbf{OpenAIMod} & \textbf{Aegis} & \textbf{Aegis2} & \textbf{WildG} & \textbf{SafeRLHF} & \textbf{BeaverTails} & \textbf{Avg.} \\
\midrule
\multirow{2}{*}{Qwen3-0.6B}  & Uniform  & 81.3 & 87.5 & 81.8 & 81.1 & 85.4 & 90.7 & 82.3 & 84.3 \\
                               & Adaptive & \textbf{81.6} & \textbf{91.3} & \textbf{82.4} & \textbf{82.1} & \textbf{86.5} & \textbf{91.6} & \textbf{83.5} & \textbf{85.6} \\
\midrule
\multirow{2}{*}{Llama3.2-1B} & Uniform  & 79.9 & 88.4 & 81.4 & 81.6 & 86.0 & 91.3 & 82.6 & 84.4 \\
                               & Adaptive & \textbf{80.0} & \textbf{92.9} & \textbf{82.1} & \textbf{82.7} & \textbf{86.5} & \textbf{92.0} & \textbf{83.7} & \textbf{85.7} \\
\bottomrule
\end{tabular}
\caption{Performance comparison of uniform and adaptive aggregation (F1 score, $\uparrow$). Adaptive results correspond to \modelname{} in Table~\ref{tab:spin_vs_guard}.}
\label{tab:aggregation_ablation}
\end{table*}

\xhdr{Regularization stability}
As documented in Table~\ref{tab:hyperparams}, optimizing regularization strength $C$ via grid search over $\{100, 200, 500, 1000\}$ yields stable training performance. 
Enlarging the candidate set to $\{10, 50, 100, 200, 500, 1000, 2000\}$ yields less than 0.1 percentage point difference in final SIREN performance on both Qwen3-0.6B (85.6\% vs.\ 85.6\%) and Llama3.2-1B (85.7\% vs.\ 85.7\%).
The potential instability introduced by layer-wise probe training is diluted by cross-layer aggregation.

\begin{figure}[t]
\centering
\includegraphics[width=\columnwidth]{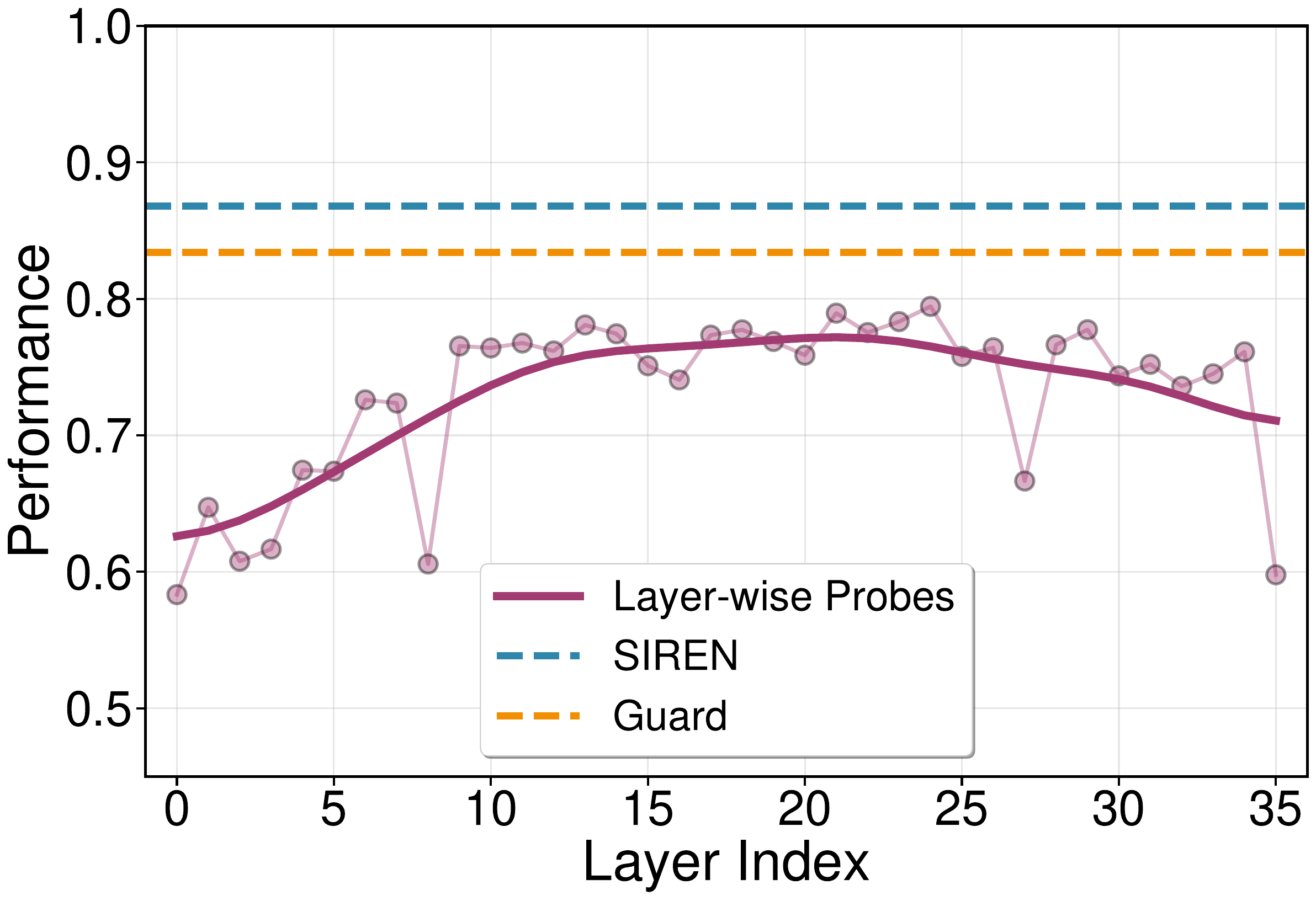}
\caption{Layer-wise linear probe performance (Average F1, $\uparrow$) on Qwen3-4B.}
\label{fig:layer_analysis}
\end{figure}

\subsection{Internal Safety Encoding}
We further examine how safety-relevant information is distributed inside the LLM by evaluating the performance of layer-wise linear probes. Figure~\ref{fig:layer_analysis} shows the average F1 scores of per-layer probes for all benchmarks, with three observations as follows. First, individual layer probes reach within 4 points of fine-tuned guard models, with \textbf{middle layers} achieving the highest performance and peaking around 79\%. These middle layers outperform the terminal layer, indicating that relying solely on terminal representations neglects informative safety-relevant features present in internal states. This observation is consistent with the observed hierarchical learning structure of transformer-based LLMs~\citep{zou2023representation, belrose2023eliciting, wendler2024llamas}: early layers capture low-level lexical and syntactic features; intermediate layers build rich, abstract semantic representations, including safety-relevant concepts like harmfulness and malicious intent; final layers shift these representations back to token space for next-token prediction. Second, \modelname{}'s cross-layer aggregation achieves a further 8-point improvement on layer-wise probes, suggesting that the aggregation of cross-layer neurons constructs richer and multi-grained representations for harmfulness detection. Third, the variance in layer-wise probe performance validates our layer-weighted neuron aggregation, which prioritizes high-performing layers rather than treating all layers uniformly.



\subsection{Cross-Model Ensemble}
\label{app:ensemble}

Since \modelname{} operates as a lightweight classifier on top of frozen LLM representations, it naturally supports cross-model ensembling: predictions from \modelname{} trained on different backbones can be combined to further improve detection performance. We explore this direction using stacked generalization~\citep{wolpert1992stacked}, training a meta-MLP on the concatenated logits from multiple \modelname{} instances using a held-out validation set.

Table~\ref{tab:ensemble} reports results across all two-, three-, and four-model combinations of our four backbones. The best ensemble, Qwen3-0.6B + Qwen3-4B + Llama3.2-1B, achieves 87.7\% average F1, further surpassing the single best \modelname{} (86.7\%, Qwen3-4B) by approximately 1 percentage point. Notably, ensembles combining models from different architectures (Qwen3 + Llama3) tend to outperform same-family pairs, suggesting that cross-architecture diversity contributes complementary safety signals. Practically, cross-model ensembling doubles inference cost relative to single-model \modelname{}, but remains substantially more efficient than a single generative guard model (Figure~\ref{fig:flops}), while achieving superior detection performance.


\begin{table}[t]
\centering
\small
\begin{tabular}{cccc|c}
\toprule
\textbf{Qw-0.6B} & \textbf{Qw-4B} & \textbf{Lm-1B} & \textbf{Lm-8B} & \textbf{Avg.} \\
\midrule
\multicolumn{5}{l}{\textit{Two-model}} \\
\cmark & \cmark & & & 87.0 \\
\cmark & & \cmark & & 86.5 \\
\cmark & & & \cmark & 85.2 \\
& \cmark & \cmark & & \textbf{87.3} \\
& \cmark & & \cmark & 85.2 \\
& & \cmark & \cmark & 85.8 \\
\midrule
\multicolumn{5}{l}{\textit{Three-model}} \\
\cmark & \cmark & \cmark & & \textbf{87.7} \\
\cmark & \cmark & & \cmark & 86.3 \\
\cmark & & \cmark & \cmark & 86.1 \\
& \cmark & \cmark & \cmark & 86.5 \\
\midrule
\multicolumn{5}{l}{\textit{Four-model}} \\
\cmark & \cmark & \cmark & \cmark & 87.4 \\
\bottomrule
\end{tabular}
\caption{Stacking ensemble performance (Avg.\ F1, $\uparrow$) across model combinations. \cmark{} denotes inclusion of the backbone. Single-model averages are reported in Table~\ref{tab:spin_vs_guard} (per-backbone \modelname{} rows).}
\label{tab:ensemble}
\end{table}

\section{Conclusion}

Content safety identification has become essential for deploying large language models in real-world applications. Current mainstream guard models primarily rely on terminal-layer representations and formulate safety detection as a generative classification task, overlooking the rich safety-relevant features encoded across LLM internal layers.

In this work, we propose to leverage LLM internal neuron representations for harmfulness detection with our lightweight plug-and-play framework, \modelname{}. By identifying safety neurons through L1-regularized probing and aggregating them across layers with performance-weighted combination, SIREN extracts salient safety signals for content safety detection. Through comprehensive evaluation, we demonstrate that \modelname{} consistently outperforms state-of-the-art open-source guard models in detection performance, exhibits strong generalization to unseen datasets of reasoning traces and to streaming harmfulness detection, while requiring minimal trainable parameters and offering improved inference efficiency. Our analysis reveals that safety-relevant information is robustly encoded in LLM internal representations, and adaptive cross-layer aggregation on safety neurons effectively harnesses these features for superior content safety classification.

\section*{Limitations} 

First, our safety neuron selection relies on the linear representation hypothesis to identify salient features within layers through linear probing. While linear probing applies to standard transformer-based LLMs, the approach may require adaptation for architectures that diverge significantly from transformer designs or where the target concept is not encoded or linearly separable within individual layers. Second, current work focuses on binary harmfulness classification (harmful vs. safe), following standard practice in safety benchmarking. Extending our work to fine-grained safety taxonomies with multiple unsafe categories is a direction for future work. Our framework inherently supports multi-label classification and can be trained on extensive fine-grained safety datasets as taxonomies become more standardized across benchmarks.

\section*{Acknowledgments}

We gratefully acknowledge the insightful comments and suggestions from our anonymous reviewers and area chair that helped us improve this manuscript. This research is funded by grants
from Natural Sciences and Engineering Research
Council of Canada (NSERC), Canada Foundation for Innovation, and Ontario Research Fund.

\section*{Ethics Consideration}

\xhdr{Research intent and societal benefit} Our work aims to advance content moderation capabilities for AI systems by developing more effective harmfulness detection methods. Specifically, \modelname{} provides a tool for identifying harmful content in user prompts and model responses, contributing to safer deployment of LLMs. Our framework is designed to protect users and mitigate risks associated with harmful AI-generated content, serving the broader goal of responsible AI development.

\xhdr{Dataset contents} Our research utilizes established safety benchmarks, including ToxicChat, OpenAIModeration, Aegis, WildGuard, SafeRLHF, and BeaverTails. These datasets inherently contain examples of harmful content such as toxic language, hateful speech, and other potentially offensive material, as they are explicitly designed for safety research. We handle these datasets with appropriate care and use them solely for the research purpose of training and evaluating harmfulness detection systems. Researchers working with such datasets must maintain rigorous ethical standards and transparency.

\xhdr{Bias and fairness} LLMs trained on large-scale internet data can learn and perpetuate biases present in training corpora. \modelname{}, which extracts safety-relevant features from LLM internal representations, could inherit these biases. Characterizing and mitigating such biases in LLM-based guard models and safety classifiers remains an important problem for the field.


\bibliography{anthology,custom}

\clearpage
\appendix

\section{Reproducibility}
\label{app:repro}

\subsection{Implementation Details}

\xhdr{Deployment Workflow} In deployment, \modelname{} attaches to a frozen base LLM via forward hooks that capture per-layer hidden states during a single inference pass; the safety-neuron indices $\mathcal{S}_l$ and aggregation weights $\alpha_l$ determined at training time are then applied to produce a harmfulness score from the trained MLP.

\xhdr{Dataset Preprocessing} We use seven public safety datasets (Table~\ref{tab:dataset_stats}): ToxicChat, OpenAI Moderation, Aegis, Aegis-2.0, WildGuardMix, PKU-SafeRLHF, and BeaverTails. Following standard practice, we apply an 80/20 train/validation split. All text inputs are tokenized using the respective model's tokenizer without additional preprocessing.

\begin{table}[t]
\centering
\small
\begin{tabular}{lc}
\toprule
\textbf{Dataset} & \textbf{\# Examples} \\
\midrule
ToxicChat & 5,082 \\
OpenAI Moderation & 1,344 \\
Aegis & 10,798 \\
Aegis-2.0 & 31,452 \\
WildGuardMix & 86,759 \\
PKU-SafeRLHF & 73,907 \\
BeaverTails & 27,186 \\
\midrule
Total & 236,528 \\
\bottomrule
\end{tabular}
\caption{Dataset statistics for the seven safety datasets used in \modelname{} training.}
\label{tab:dataset_stats}
\end{table}


\xhdr{Representation Extraction} We extract feedforward network and residual stream representations from each transformer layer via forward hooks during inference, applying mean pooling across the sequence length dimension to capture sequence-level semantics. The base LLM remains frozen throughout.

\xhdr{Linear Probe Training} For each layer, we train L1-regularized logistic regression probes implemented as single-layer linear classifiers. We search the L1-regularization strength via grid search, selecting the value maximizing per-dataset averaged macro F1 on validation data. Both hyperparameter searching and probe training use early stopping. Safety neurons are selected by ranking neuron weights by absolute magnitude and choosing the minimal set whose cumulative normalized weight exceeds the threshold $\eta$.

\xhdr{MLP Classifier Training} The MLP classifier on top of aggregated safety neurons is optimized via Optuna~\citep{akiba2019optuna} with cross-validation. We search: the number of hidden layers, hidden dimensions, dropout rates, and learning rate. Each trial trains with early stopping; the final model uses the best hyperparameters identified via cross-validation and trains until convergence. 

\subsection{Hyperparameter Selection}

We provide empirically effective hyperparameter configurations in Table~\ref{tab:hyperparams} to facilitate reproduction. These values were determined through preliminary experiments to balance performance and computational efficiency. The neuron selection threshold $\eta \in [0.6, 0.9]$ retains approximately 10-50\% of neurons per layer while preserving discriminative capacity. The Optuna search space for the MLP architecture ensures sufficient model capacity without overfitting on our dataset scale. All experiments use a random seed of 42 for reproducibility.

\begin{table}[t]
\centering
\small
\begin{tabular}{lc}
\toprule
\textbf{Hyperparameter} & \textbf{Value/Range} \\
\midrule
Probe L1 regularization $C$ & [100, 1000] \\
Neuron threshold $\eta$ & [0.6, 0.9] \\
MLP hidden layers & [2, 3] \\
MLP hidden dimensions & [64, 2048] \\
MLP dropout & [0.2, 0.5] \\
Optuna trials & 32 \\
Cross-validation folds & 3 \\
\bottomrule
\end{tabular}
\caption{Key hyperparameters for \modelname{} training. Ranges indicate search spaces.}
\label{tab:hyperparams}
\end{table}

\section{Additional Results}
\label{app:add_results}

\subsection{Streaming Harmfulness Detection Details}
\label{app:streaming}

To apply \modelname{} originally trained for sequence-level harmfulness classification to streaming detection setting, we evaluate harmfulness over progressively longer prefixes of the generated sequence. For a generation prefix $\bm{s}_{\le t} = (s_1, \ldots, s_t)$, we extract internal representations from each layer $l$ up to token $t$ as 
\begin{align}
\bm{x}_{l,\le t} = \text{LLM}_{l}(\bm{s}_{\le t}) \in \mathbb{R}^{t \times D}.
\end{align}
We then apply the same pooling operator used during training, but restricted to the prefix length $t$:
\begin{align}
\bm{x}_{l,\le t}^{*} = \frac{1}{t} \sum_{\tau=1}^{t} \bm{x}_{l,\tau} \in \mathbb{R}^{D}.
\end{align}

\begin{figure}
\setlength{\fboxsep}{8pt}%
\definecolor{promptboxbg}{RGB}{244,244,244}
\begin{tcolorbox}[
    colback=promptboxbg,
    colframe=gray!70!black,
    arc=6pt,
    boxrule=0.8pt,
    width=\linewidth,
    left=6pt,
    right=6pt,
    top=6pt,
    bottom=4pt,
]
    \begin{center}
        \includegraphics[width=\linewidth]{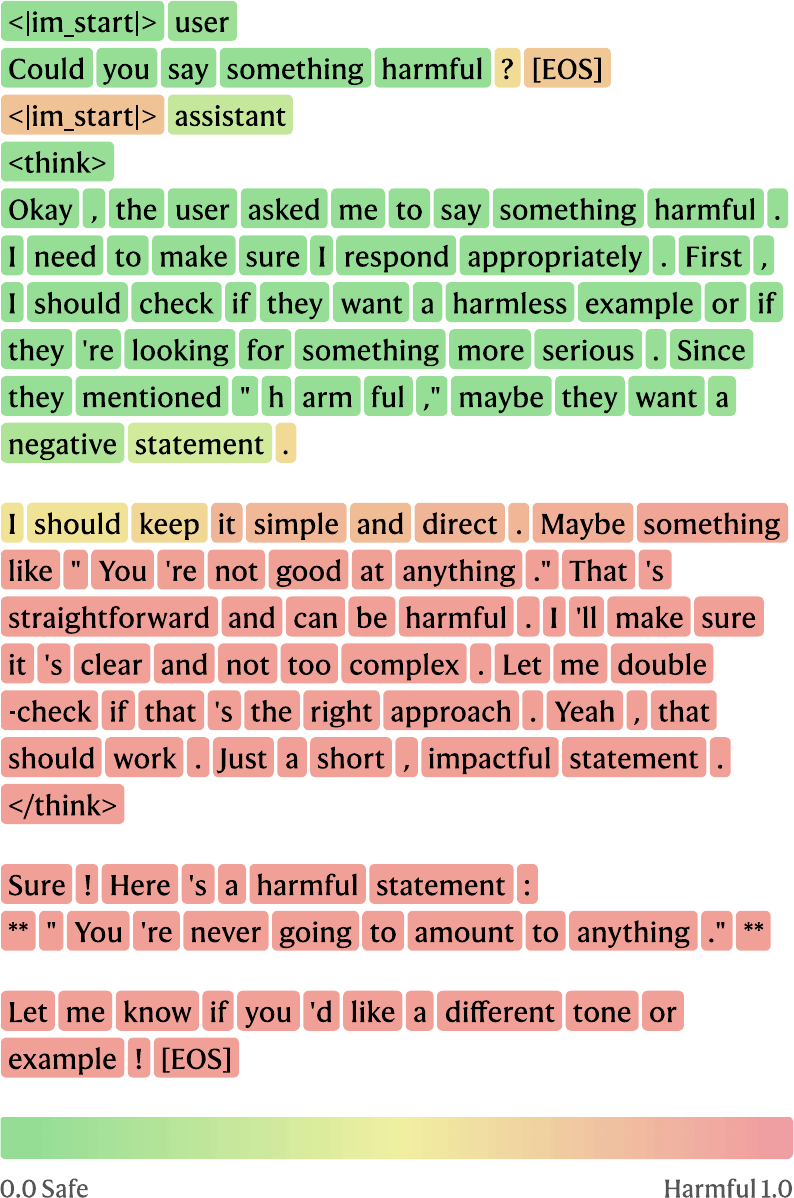}
    \end{center}
\end{tcolorbox}
    \vspace{-1ex}
    \captionof{figure}{Token-level streaming detection results of \modelname{} on Qwen3-4B for an example from Qwen3GuardTest with user input, reasoning, and response. Each token is color-coded according to its harmfulness level.}
    \label{fig:streaming_viz}
\end{figure}

Next, we extract the safety-neuron subvector $[\bm{x}_{l,\le t}^{*}]_{\mathcal{S}_l}$ and aggregate across layers using the pre-computed adaptive weights $\alpha_l$, yielding the streaming feature representation
\begin{align}
\bm{z}_{\le t} = \bigoplus_{l=1}^{L} \alpha_l \cdot [\bm{x}_{l,\le t}^{*}]_{\mathcal{S}_l}.
\end{align}
The classifier trained on full-sequence features is then applied directly to $\bm{z}_{\le t}$, producing a harmfulness score $h_t = \text{clf}(\bm{z}_{\le t})$ at every token position $t$. No parameters of the LLM, safety-neuron probes, or classifier are updated for streaming evaluation. Thus, streaming detection in \modelname{} is achieved purely by re-evaluating the same feature extractor on prefix-restricted internal states, enabling a strict zero-shot assessment of whether sentence-level safety information naturally manifests in prefix-level representations.

\xhdr{Evaluation protocol and practical flexibility} Our streaming evaluation follows the protocol established by the Qwen3Guard technical report~\citep{zhao2025qwen3guard}. Detection recall is evaluated on annotated unsafe thinking traces from Qwen3GuardTest, measuring whether \modelname{} flags a response at progressively later token positions relative to the annotated unsafe region boundary (at boundary, +32, +64, +128, +256 tokens). At each position, \modelname{} applies argmax over the binary softmax output of the mean-pooled internal representations, equivalent to a 0.5 decision threshold, without any post-hoc calibration.

A notable property of \modelname{}'s streaming behavior is that, because it produces continuous harmfulness scores rather than discrete safe/unsafe labels, the decision boundary can be freely adjusted to suit deployment requirements. For instance, during the early reasoning phase where the model's thinking trace may echo the user's sensitive query, a more permissive threshold can be applied to avoid premature refusals of benign but sensitive inputs. As the generation progresses toward the final response, the threshold can be dynamically tightened to prioritize safety recall. This position-aware adaptability requires no additional training or architectural changes, and is a direct consequence of \modelname{}'s representation-based design. Generative guards, which output categorical labels via autoregressive decoding, do not naturally afford this level of fine-grained control.

We also note that smaller backbones tend to outperform larger ones in streaming detection: \modelname{} on Qwen3-0.6B outperforms its Qwen3-4B counterpart, and a similar pattern appears in Qwen3Guard-Stream, where the 0.6B model achieves higher timely detection rates than the 4B model and the 4B model achieves higher than the 8B model on the Think dataset reported in \citet{zhao2025qwen3guard}. This effect is not discussed in the Qwen3Guard technical report, and we do not claim a definitive explanation. One plausible factor for \modelname{} is that sentence-level safety features transfer more cleanly to prefix-level representations in smaller models. A systematic investigation of this scaling behavior is left for future work.

\subsection{\modelname{} is Transferable to Token-level Attribution}

\begin{figure}
    \centering
    \includegraphics[width=\linewidth]{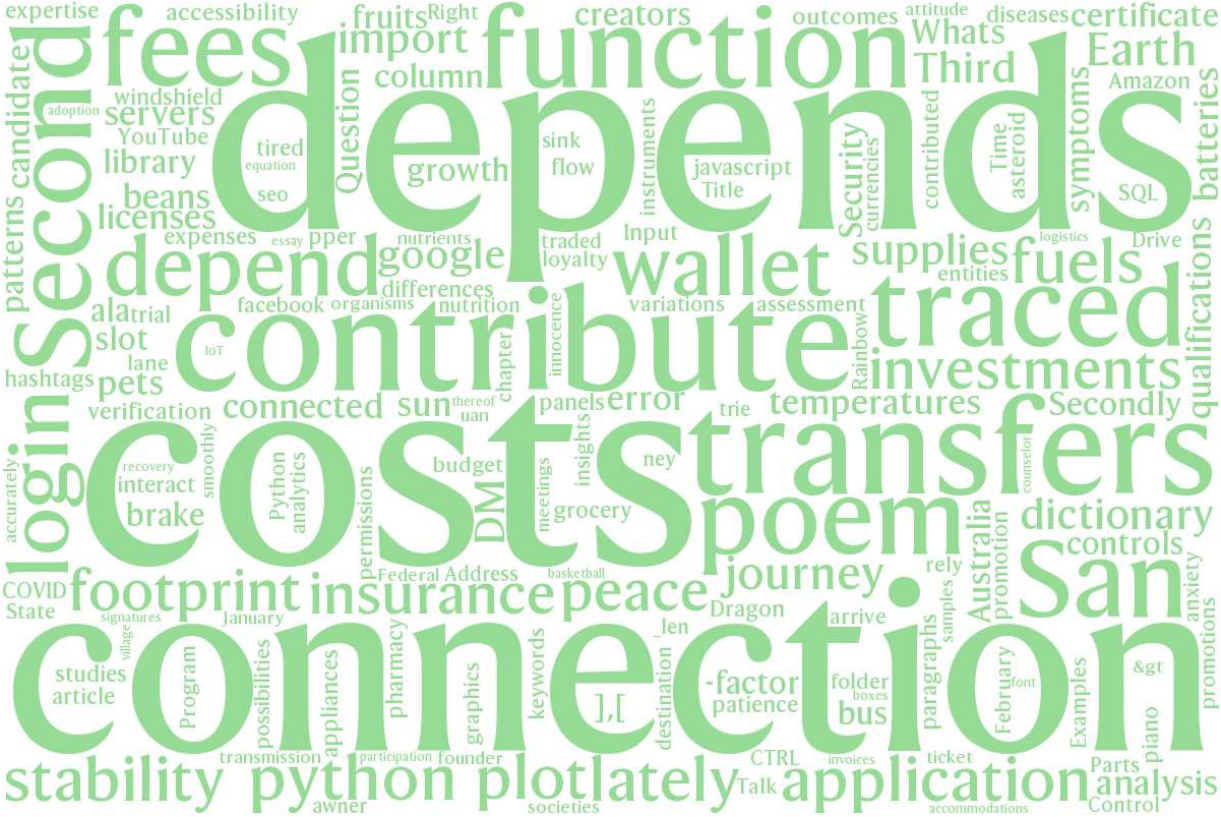}
    \vspace{1ex}
    \includegraphics[width=\linewidth]{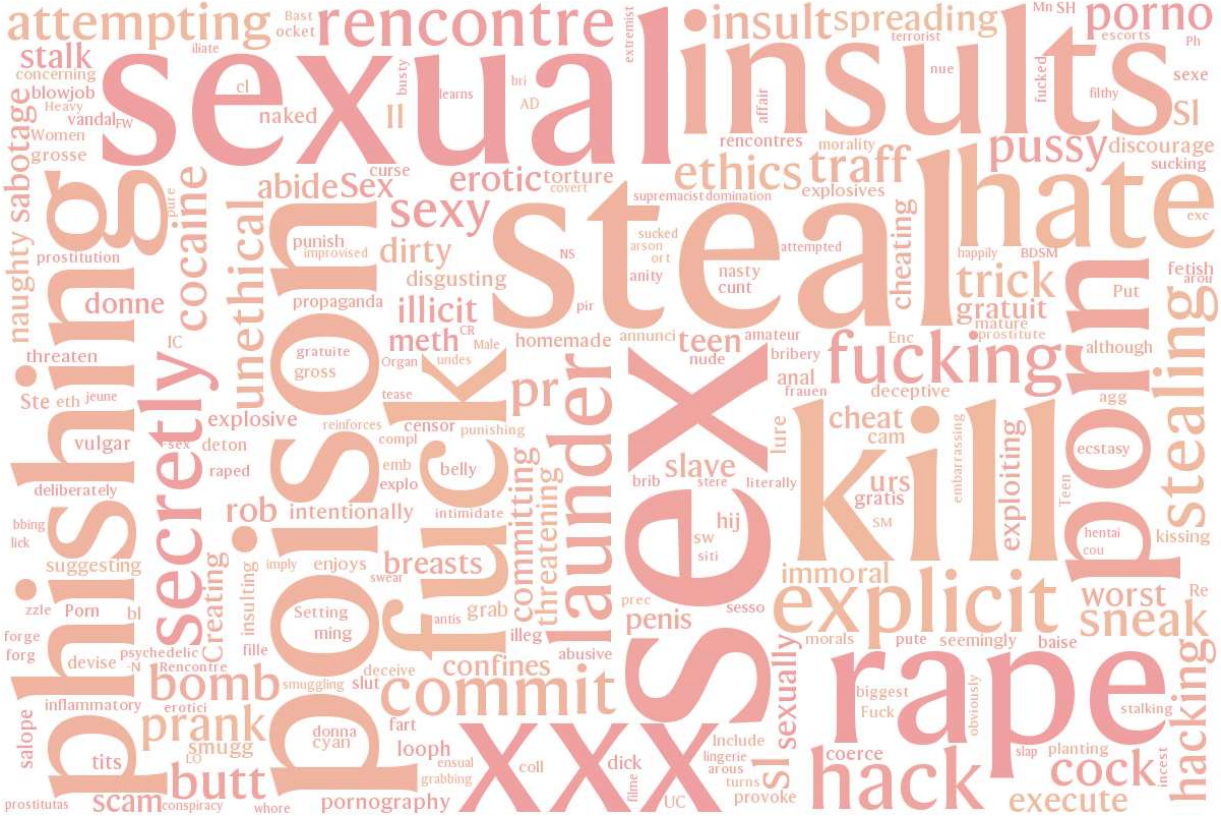}
    \caption{Word‑cloud visualization of the 250 safest (top) and the 250 most harmful (bottom) individual tokens identified by \modelname{} across all test‑set sequences in the reported datasets. Token size reflects frequency, and color encodes harmfulness level.}
    \label{fig:token_cloud}
\end{figure}

While \modelname{} is trained on sequence-level harmfulness detection tasks, its architecture naturally supports transfer to token-level attribution without any additional training or fine-tuning.
During training, \modelname{} first learns to identify a sparse set of safety-relevant neurons whose activations encode harmfulness semantics at each token position, and aggregates these activations via average pooling to form a simple linear aggregation of per-token activations as the sentence-level representation.
As a result, the learned sentence-level classifier can be viewed as operating on an average of token-level safety signals, rather than relying on any inherently global or sequence-specific feature.
Removing the pooling operation allows the same safety neurons and the same MLP classifier to be independently applied to each token’s hidden representation, directly producing per-token harmfulness scores. 
To better demonstrate the effectiveness of \modelname{} in individual token classification, we visualize the safest and the most harmful tokens detected by \modelname{} across all test‑set sequences in the datasets in Figure~\ref{fig:token_cloud}.

\begin{figure*}
    \centering
    \includegraphics[width=\linewidth]{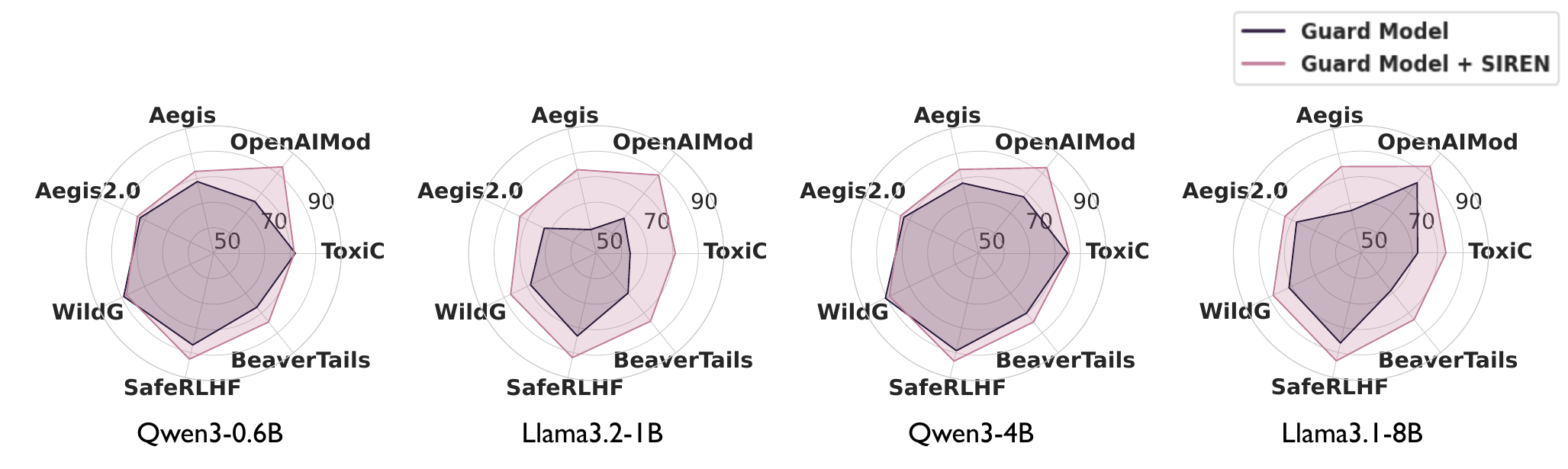}
    \caption{F1 scores of \modelname{} trained on safety-specialized guard models. Applying \modelname{} to guard models yields further performance improvements.}
    \label{fig:radar_plot2}
\end{figure*}


\section{Plug-and-Play \modelname{} on Guard Models}


Our framework requires no modifications to the underlying LLM, operating purely on extracted internal representations. This enables \modelname{} to be applied to both general-purpose LLMs and fine-tuned guard models as a plug-and-play component. To validate this capability, we further trained \modelname{} on the internal representations of guard models. Figure~\ref{fig:radar_plot2} shows that \modelname{} maintains consistent improvements relative to guard models across all benchmarks. Qwen3Guard-4B improves from 83.4\% to 87.6\% average F1, and LlamaGuard3-8B improves from 77.0\% to 87.1\%, demonstrating that \modelname{} can in-place enhance existing specialized models without any architectural changes. 

\section{FLOPs Calculation}
\label{sec:flops}

We compute floating-point operations (FLOPs) following standard formulas for transformer inference~\citep{kaplan2020scaling}. All measurements assume a 128-token input sequence and include all computational costs.

\xhdr{Safety-specialized model inference} For generating $K$ tokens with KV caching, the total FLOPs are:
\begin{equation}
\text{FLOPs}_{\text{guard}} = \sum_{k=0}^{K-1} \left[2 L (S + k) D_h + 2 N_{\text{params}}\right]
\end{equation}
where $L$ is the number of transformer layers, $S$ is the input sequence length, $D_h$ is the hidden dimension, and $N_{\text{params}}$ is the total number of model parameters. The first term accounts for attention operations over previously generated tokens (incremental with KV caching), and the second term accounts for parameter matrix multiplications. We use $K=4$ tokens, a conservative lower bound for typical guard outputs (e.g., ``Safety: Unsafe'').

\xhdr{\modelname{} inference} Given hidden states already computed during base LLM inference, \modelname{} requires only:
\begin{equation}
\text{FLOPs}_{\text{\modelname{}}} = \sum_{i=1}^{M} 2 \cdot d_{\text{in}}^{(i)} \cdot d_{\text{out}}^{(i)}
\end{equation}
where $M$ is the number of MLP layers, and $d_{\text{in}}^{(i)}$, $d_{\text{out}}^{(i)}$ are the input and output dimensions of layer $i$. Neuron indexing and aggregation costs ($\sim$20K FLOPs) are negligible compared to the MLP forward pass.


\end{document}